\documentclass[final,authoryear,3p,times]{elsarticle}

\usepackage{amssymb}
\usepackage[T1]{fontenc}
\usepackage{amsmath, amssymb, amsthm}

\newtheorem{theorem}{Theorem}[section]
\newtheorem{corollary}{Corollary}[section]
\newtheorem{lemma}{Lemma}[section]
\newtheorem{proposition}{Proposition}[section]

\def\e{\epsilon}
\def\X{{\cal X}_o}

\journal{Statistics and Probability letters}

\begin{document}

\begin{frontmatter}

\title{On approximation of smoothing probabilities for hidden Markov models}


\author{Jüri Lember\fnref{grant}}
\fntext[grant]{Estonian science foundation grant no 7553}

\address{Tartu University, J. Liivi 2 - 507, Tartu 50408, Estonia}
\ead{jyril@ut.ee}

\begin{abstract}
We consider the smoothing probabilities of hidden Markov model
(HMM). We show that under fairly general conditions for HMM, the
exponential forgetting still holds, and the  smoothing probabilities
can be well approximated with the ones of double sided HMM. This
makes it possible to use ergodic theorems. As an applications we
consider the pointwise maximum a posteriori segmentation, and show
that the corresponding risks converge
\end{abstract}


\end{frontmatter}


\section{Introduction}
\label{Intro} \noindent Let $Y=\{Y_k \}_{-\infty}^{\infty}$ be a
double-sided stationary MC with states $S=\{1,\ldots,K\}$ and
irreducible aperiodic transition matrix $\big(P(i,j)\big)$. Let
$X=\{X_k\}_{-\infty}^{\infty}$ be the (double-sided) process such
that: 1) given $\{Y_k\}$  the random variables $\{X_k\}$ are
conditionally independent; 2) the distribution of $X_n$ depends on
$\{Y_k \}$ only through $Y_n$. The process $X$ is sometimes called
the {\it hidden Markov process} (HMP) and the pair $(Y,X)$ is
referred to as {\it hidden Markov model} (HMM). The name is
motivated by the assumption that the process $Y$ (sometimes called
as {\it regime}) is non-observable. The distributions $P_i:=P(X_1\in
\cdot|Y_1=i)$ are called {\it emission distributions}. We shall
assume that the emission distributions are defined on a measurable
space  $({\cal X},{\cal B})$, where ${\cal X}$ is usually
$\mathbb{R}^d$ and ${\cal B}$ is the Borel $\sigma$-algebra. Without
loss of generality, we shall assume that the measures $P_i$ have
densities $f_i$ with respect to some reference measure $\mu$. Since
our study is mainly motivated by statistical learning, we would like
to be consistent with the notation used there ($X$ for the
observations and $Y$ for the latent variables) and therefore our
notation differs from the one used in the HMM literature, where
usually $X$ stands for the regime and $Y$ for the observations.
\\
HMM's are widely used in various fields of applications, including
speech recognition \citep{rabiner, jelinek}, bioinformatics
\citep{koski, BioHMM2}, language processing, image analysis
\citep{gray2} and many others. For general overview about HMM's, we
refer to textbook \citep{HMMraamat} and overview paper \citep{HMP}.\\
The central objects of the present papers are the {\it smoothing
probabilities} $P(Y_t=s|X_z,\ldots,X_n),$ where $t,z,n\in
\mathbb{Z}$ and $s\in S$. They are important tools for making the
inferences about the regime at time $t$. By Levy's martingale
convergence theorem, it immediately follows that as $n\to \infty$,
\begin{equation}\label{levy} P(Y_t=s|X_z,\ldots,X_n)\to
P(Y_t=s|X_z,\ldots)=:P(Y_t=s|X^{\infty}_z),\quad
\rm{a.s.}.\end{equation}
Let $P(Y_t\in \cdot|X_z^{\infty})$ denote the  $K$-dimensional
vector of probabilities from the right side of (\ref{levy}). By
martingale convergence theorem, again, as $z\to -\infty$
$$P(Y_t=s|X^{\infty}_{z})\to
P(Y_t=s|\cdots,X_{-1},X_0,X_1,\ldots)=:
P(Y_t=s|X_{-\infty}^{\infty}),\quad\text{a.s.}.$$ The double-sided
{\it smoothing process} $\{P(Y_t\in
\cdot|X_{-\infty}^{\infty})\}_{-\infty}^{\infty}$ is stationary and
ergodic, hence for this process the ergodic theorems hold. To be
able to us these ergodic theorems for establishing the limit
theorems in terms of smoothing probabilities
$P(Y_t=s|X_z,\ldots,X_n),$ it is necessary to approximate it with
double-sided smoothing process. This approach is, among others, used
in \citep{Bickel}. In other words, we are interested in bounding the
difference $\|P(Y_t\in \cdot|X_z^{n})-P(Y_t\in
\cdot|X_{-\infty}^{\infty})\|$, where $\|\cdot \|$ stands for total
variation distance. Our first main result, Corollary \ref{cor1}
states that under so-called cluster assumption {\bf A}, there exists
a bounded random variable $C$ and a constant $\rho_o\in (0,1)$ such
that for every $z_2, z_1, t, n$ such that $z_2\leq z_1\leq 1\leq
t\leq n$,
\begin{equation*}
\|P\big(Y_{t}\in \cdot|X_{z_1}^n\big)-P\big(Y_{t}\in \cdot
|X_{z_2}^n\big)\|\leq C_1{\rho_1}^{t-1},\quad \rm {a.s.}.
\end{equation*}
 Similar results can
be found in the literature for the special case, where transition
matrix has all positive entries or the emission densities $f_i$ are
all positive \citep{LeGlandMevel, ungarlased, HMMraamat}. Both
conditions are restrictive, and the assumption {\bf A} relaxes
them.\\
We go one more step further, considering the approximation of
smoothing probabilities with two-sided limits. Our second main
result, Theorem \ref{main} states that under {\bf A}, for every
$z\leq 1\leq t\leq k\leq n$
\begin{equation*}\label{thm}
 \|P(Y_t\in
\cdot|X_z^{n})-P(Y_t\in \cdot|X_{-\infty}^{\infty})\|\leq
C_1\rho_o^{t-1}+C'_k\rho_o^{k-t}\quad{\rm a.s.},\end{equation*}
where $\rho_o\in (0,1)$ is a fixed constant, $C_1$ is a finite
random variable as in the previous bound and $\{C'_k \}$ is a finite
ergodic process. Of course, without  the ergodic property, the
existence of $\{C'_k\}$ would be trivial, however, as shown in the
proof of Theorem \ref{riskthm}, the ergodic property makes the bound
useful in applications.\\
The condition {\bf A} was introduced  in \citep{IEEE, AVT4}, where
under the conditions slightly stronger than {\bf A} the existence of
infinite Viterbi alignment was shown. The technique used in these
papers differs heavily rom the one in the present paper, yet the
same assumption appears. This implies that {\bf A} is indeed
essential
for HMM's. \\
Our motivation in studying  the limit theorems of smoothing
processes comes from the segmentation theory. Generally speaking,
the {\it segmentation} problem consists of estimating the unobserved
realization of underlying Markov chain $Y_1,\ldots, Y_n$ given the
$n$ observations from from HMP $X_1,\ldots,X_n$: $x^n:=x_1,\ldots
,x_n$. Formally, we are looking for a mapping $g: {\cal X}^n \to
S^n$ called {\it classifier} that maps every sequence of
observations into a state sequence, see \citep{seg, kuljus} for
details. For finding the best $g$, it is natural to associate to
every state sequence $s^n\in S^n$ a measure of goodness
 $R(s^n|x^n)$, referred to as the
{\it risk of $s^n$}. The solution of the segmentation problem is
then the state sequence with minimal risk. In the framework of
pattern recognition theory, the risk is specified via {\it
loss-function} $L: S^n\times S^n \to [0,\infty]$, where $L(a^n,b^n)$
measures the loss when the actual state sequence is $a^n$ and the
prognosis is $b^n$. For any state sequence $s^n\in S^n$, the  risk
is then
\begin{equation}\label{risk}
R(s^n|x^n):=E[L(Y^n,s^n)|X^n=x^n]=\sum_{a^n\in
S^n}L(a^n,s^n)P(Y^n=a^n|X^n=x^n).
\end{equation}
In this paper, we consider the case when the loss function is given
as
\begin{equation}\label{point-loss}L(a^n,b^n)={1\over
n}\sum_{i=1}^nl(a_i,b_i),\end{equation} where $l(a_i,b_i)\geq 0$ is
the loss of classifying the $i$-th symbol $a_i$ as $b_i$. Typically,
for every state $s$, $l(s,s)=0$. Most frequently, $l(s,s')=I_{\{s\ne
s'\}}$ and then the risk $R(s^n|x^n)$ just counts the expected
number of misclassified
symbols.\\
Given a classifier $g$, the quantity $R(g,x^n):=R(g(x^n)|x^n)$
measures the goodness of it when applied to the observations $x^n$.
When $g$ is optimal in the sense of risk, then
$R(g,x)=\min_{s^n}R(s^n|x^n)=:R(x^n)$. We are interested in the
random variables $R(g,X^n)$. When $g$ is maximum likelihood
classifier -- so called Viterbi alignment -- and HMM satisfies {\bf
A}, then (under an additional mild assumption),  it can be shown
that  there exists a constant $R_v$ such that $R(g,X^n)\to R_v$ a.s.
\citep{caliebe2,IEEE}. In this paper, we show that under {\bf A},
the similar results holds for optimal alignment: there exists a
constant $R$ such that $R(X^n)\to R$, a.s.. Those numbers (clearly
$R_v\geq R$) depend only on the model and they measure the
asymptotic goodness of the segmentation. If $l(s,s')=I_{\{s\ne
s'\}}$, then $R_v$ and $R$ are the asymptotic symbol-by symbol
misclassification rates when Viterbi alignment or the best alignment
(in given sense) are used in segmentation, respectively.
\section{Approximation of the smoothing probabilities}
\subsection{Preliminiaries}
\noindent
Throughout the paper, let $x_u^v$ where $u,v\in \mathbb{Z}$, $u\leq
v$ be a realization of $X_u,\ldots,X_v$. We refer to $x_u^v$ as the
{\it observations}. When $u=1$, then it is omitted from the
notation, i.e. $x^n:=x_1^n$. Let $p(x_u^v)$ stand for the likelihood
of the observations $x_u^v$. For every $u \leq t \leq v$ and $s\in
S$, we also define the forward and backward variables
$\alpha(x_u^t,s)$ and $\beta(x^v_{t+1}|s)$ as follows
$$\alpha(x^t,s):=p(x_u^t|Y_t=s)P(Y_t=s),\quad \beta(x^v_{t+1}|s):=\left\{
                                                                    \begin{array}{ll}
                                                                      1, & \hbox{if $t=v$;} \\
                                                                      p(x_{t+1}^v|Y_t=s), & \hbox{if $t<v$.}
                                                                    \end{array}
                                                                  \right.
.$$ Here $p(x_u^t|Y_t=s)$ and $p(x_{t+1}^v|Y_t=s)$ are conditional
densities  \citep[see also][]{HMP}. The backward variables can be
calculated recursively (backward recursion):
$$\beta(x_{u+1}^n|s)=\sum_{i\in
S}P(s,i)f_i(x_{u+1})\beta(x_{u+2}^n|i).$$ For every $t\in
\mathbb{Z}$, we shall denote by $\pi_t[x_u^v]$ the $K$-dimensional
vector of conditional probabilities $P(Y_t\in \cdot|X_u^v=x_u^v)$.
Our first goal is to bound the difference
$\pi_t[x_{z_1}^n]-\pi_t[x_{z_2}^n]$, where $z_2\leq z_1\leq 1\leq
t\leq n$. For that, we shall follow the approach in
\citep{HMMraamat}. It bases on the observation that given the
observations $x_{z}^n$, the underlying chain $Y_{z},\ldots,Y_n$ is a
conditionally inhomogeneous MC, i.e for every $z\leq k < n$ and
$j\in S$
\begin{align*}
P(Y_{k+1}=j|Y_{z}^{k}=y_{z}^{k},X_{z}^n=x_{z}^n)=P(Y_{k+1}=j|Y_{k}=y_{k},X_{z}^n=x_{z}^n)=:F_k(y_k,j),
\end{align*}
where for every $i\in S$, $F(i,j)$ is called the {\it forward
smoothing probability} \citep[Prop. 3.3.2]{HMMraamat}, also
\citep[(5.2.1)]{HMP}. It is known \citep[(5.21)]{HMMraamat},
\citep[(3.30)]{HMP} that if $\beta(x_{k+1}^n|i)>0$, then
\begin{equation}\label{Fdef}
F_k(i,j)={P(i,j)f_j(x_{k+1})\beta(x_{k+2}^n|j)\over
\beta(x_{k+1}^n|i)}.
\end{equation}
When $\beta(x_{k+1}^n|i)=0$, we define $F_k(i,j)=0$ $\forall j\in
S$. Note that the matrix $F_k$ depends on the observations
$x_{k+1}^n$, only. This dependence is sometimes denoted by
$F_k[x_{k+1}^n]$. With the matrices $F_k$, for every $t$ such that
$z\leq t\leq n$, it holds \citep[e.g.][(4.30)]{HMMraamat}
\begin{equation}\label{pi}
\pi'_t[x^n_{z}]=
\pi'_{z}[x_{z}^n]\big(\prod_{i=z}^{t-1}F_i[x_{i+1}^n]\big),
\end{equation}
where $'$ stands for transposition.  For $n\geq t\geq 1\geq z_1\geq
z_2$, thus
\begin{equation}\label{moodud}
\big(\pi_t[x^n_{z_1}]-\pi_{t}[x_{z_2}^n]\big)'=\big(\pi_1[x_{z_1}^n]-\pi_1[x_{z_2}^n]\big)'\prod_{i=1}^{t-1}F_i[x_{i+1}^n].
\end{equation}
Let $\pi_1$ and $\pi_2$ be  two probability measures on $S$. If  $A$
is  a transition matrix on $S$, then $A'\pi_i$, $(i=1,2)$ is a
vector that corresponds to a probability measure. We are interested
in total variation distance between the measures $A'\pi_1$ and
$A'\pi_2$.  The approach in this paper uses the fact that the
difference between measures can be bounded as follows \citep[Cor.
4.3.9]{HMMraamat}
\begin{equation}\label{hinnang}
\|A'\pi_1-A'\pi_2\|=\|A'(\pi_1-\pi_2)\|\leq
\|\pi_1-\pi_2\|\delta(A),
\end{equation}
where $\delta(A)$ is {\it Dobrushin coefficient} of $A$ defined as
follows
\begin{equation*}
\delta(A):={1\over 2}\sup_{i,j\in S}\|A(i,\cdot)-A(j,\cdot)\|.
\end{equation*}
Here, $A(i,\cdot)$ is the $i$-th row of the matrix. Hence, applying
(\ref{hinnang}) to (\ref{moodud}), we get \citep[Prop.
4.3.20]{HMMraamat}
\begin{equation}\label{hinnang2}
\big\|\pi_t[x^n_{z_1}]-\pi_{t}[x_{z_2}^n]\big\|\leq
\big\|\pi_1[x_{z_1}^n]-\pi_1[x_{z_2}^n]\big\|\delta\Big(\prod_{i=1}^{t-1}F_i[x_{i+1}^n]\Big)\leq
2\delta\Big(\prod_{i=1}^{t-1}F_i[x_{i+1}^n]\Big).
\end{equation}
Another useful fact is that for two transition matrices $A,B$, it
holds \citep[see, e.g.][Prop 4.3.10]{HMMraamat} $\delta(AB)\leq
\delta(A)\delta(B),$ hence, the right hand side of (\ref{hinnang2})
can be further bounded above with
$2\prod_{i=1}^{t-1}\delta\big(F_i[x_{i+1}^n]\big)$. The Dobrushin
coefficient of $A$ can be estimated above using so-called {\it
Doeblin condition}: If  there exists $\e>0$ and a probability
measure $\lambda=(\lambda_1,\ldots,\lambda_K)$ on $S$ such that
\begin{equation}\label{doeblin}
A(i,j)\geq \e \lambda_j,\quad \forall i,j\in S,
\end{equation}
then $\delta(A)\leq 1-\e$  \citep[Lemma 4.3.13]{HMMraamat}. This
condition holds, for example, if all entries of $A$ are positive. If
$F_i$ satisfies Doeblin conditions, then the right hand side
converges to zero exponentially with $t$.
\subsection{Cluster-assumption}
\noindent Recall that $f_i$ are the densities of $P(X_1\in
\cdot|Y_1=i)$ with respect to some reference measure $\mu$ on
$({\cal X},{\cal B})$. For each $i\in S$, let
$G_i:=\{x\in\mathcal{X}: f_i(x)>0\}.$
We call a subset $C\subset S$  {\it a cluster} if the following
conditions are satisfied:
$$\min_{j\in C}P_j(\cap _{i\in C}G_i)>0,~{\rm and}\,\max_{j\not\in C}P_j(\cap _{i\in C}G_i)=0.$$
Hence, a cluster is a maximal subset of states such that $G_C=\cap
_{i\in C}G_i$, the intersection of the supports of the corresponding
emission distributions, is  `detectable'. Distinct clusters need not
be disjoint and  a cluster can consist of a single state. In this
latter case such a state is not hidden, since it is exposed by any
observation it emits. When $K=2$, then $S$ is the only  cluster
possible, since otherwise the underlying Markov chain would cease to be hidden.\\
Let $C$ be a cluster. The existence of $C$ implies the existence of
a set $\X \subset \cap_{i\in C }G_i$ and $\e>0$, $K<\infty$
 such that $\mu (\X)>0$, and $\forall x\in
{\X}$, the following statements hold: (i) $\e<\min_{i\in C} f_i(x)$;
(ii) $\max_{i\in C} f_i(x)<K$; (iii) $\max_{j\not \in C} f_j(x)=0$.
For proof, see \citep{IEEE}.\\\\
{\bf Assumption A:} There exists a cluster $C\subset S$ such that
the sub-stochastic matrix $R=(P(i,j))_{i,j\in C}$ is primitive (i.e.
there is a positive integer $r$ such that the $r$th power $R$
 is strictly positive ).\\\\
Clearly assumption {\bf A} is satisfied, if the matrix $P$ has all
positive elements. Since any irreducible aperiodic matrix is
primitive, the assumption {\bf A} is also satisfied, if the  the
densities $f_i$ satisfy the following condition: For every $x\in
{\cal X}$, $\min_{i\in S}f_i(x)>0$, i.e. for all $i\in S$,
$G_i={\cal X}$. Thus {\bf A} is more general than the {\it strong
mixing condition} \citep[Assumption 4.3.21]{HMMraamat} and
\citep[Assumption 4.3.29]{HMMraamat}.
For more general discussion about {\bf A}, see \citep{AVT4,IEEE}.\\
In the following, we assume {\bf A}. Let $C$ be the corresponding
cluster, and let $\X$ be the corresponding set.
\begin{proposition}\label{clust-A} Let $x_{k+1}^n$ be such that $p(x_{k+1}^n)>0$ and  $x_{k+1}^{k+r}\in \X^r$. Then
\begin{equation}\label{25}
\delta\big(\prod_{i=k}^{k+r-1}F_{i}[x^n_{k+1}]\big)\leq
1-{\min_{i,j}R^r(i,j)\over \max_{i,j} R^r(i,j)}\Big({\e\over
K}\Big)^r=:\rho<1.
\end{equation}
\end{proposition}
\begin{proof}
Let $A:=\prod_{l=0}^{r-1}F_{k+l}$. Using backward recursion, it
follows that for every $r\geq 1$
\begin{align*}\label{valem}
A(i,j)={\sum_{i_1}\cdots \sum_{i_{r-1}}
P(i,i_1)f_{i_1}(x_{k+1})\ldots
f_{i_{r-1}}(x_{k+r-1})P(i_{r-1},j)f_j(x_{k+r})\beta(x_{k+r+1}^n|j)\over
\sum_j\sum_{i_1}\cdots \sum_{i_{r-1}} P(i,i_1)f_{i_1}(x_{k+1})\ldots
f_{i_{r-1}}(x_{k+r-1})P(i_{r-1},j)\beta(x_{k+r+1}^n|j)}
\end{align*}
Since  $x_{k+1}^{k+r}\in \X^r$, then by (iii), (ii) and (i), thus
for every $i,j\in S$
\begin{align*}
A(i,j)
&={\sum_{i_1\in C}\cdots \sum_{i_{r-1}\in
C}P(i,i_1)f_{i_1}(x_{k+1})\cdots
P(i_{r-1},j)f_j(x_{k+r})\beta(x^n_{k+r+1}|j)\over \sum_{j\in
C}\sum_{i_1\in C}\cdots \sum_{i_{r-1}\in
C}P(i,i_1)f_{i_1}(x_{k+1})\cdots
P(i_{r-1},j)f_j(x_{k+r})\beta(x^n_{k+r+1}|j)}\\
&\geq \Big({\e \over K}\Big)^r{\big(\sum_{i_1\in C}\cdots
\sum_{i_{r-1}\in C}P(i,i_1)\cdots
P(i_{r-1},j)\big)\beta(x^n_{k+r+1}|j)\over \sum_{j\in C} \big(
\sum_{i_1\in C}\cdots \sum_{i_{r-1}\in C}P(i,i_1)\cdots
P(i_{r-1},j)\big)
\beta(x^n_{k+r+1}|j)}\\
&= \Big({\e \over K}\Big)^r{R^r(i,j) \beta(x^n_{k+r+1}|j)\over
\sum_j R^r(i,j)\beta(x^n_{k+r+1}|j)}\geq {\min_{i,j}R^r(i,j) \over
\max_{i,j}R^r(i,j)}\Big({\e \over
K}\Big)^r{\beta(x^n_{k+r+1}|j)\over \sum_j
\beta(x^n_{k+r+1}|j)}=\e_o\lambda_j,
\end{align*}
where $$\e_o={\min_{i,j}R^r(i,j) \over
\max_{i,j}R^r(i,j)}\Big({\e\over K}\Big)^r,\quad
\lambda_j:={\beta(x^n_{l+r+1}|j)\over \sum_j
\beta(x^n_{l+r+1}|j)}.$$ Since
$$p(x_{k+1}^n)=\sum_j\alpha(x_{k+1}^{k+r},j)\beta(x^n_{k+r+1}|j)>0,$$
there must be a $j\in S$ such that $\beta(x^n_{l+r+1}|j)>0$. So
$(\lambda_j)_{j\in S}$ is a probability measure and Doeblin
condition holds.\end{proof}
\begin{lemma}\label{lemmake} Let $x_{z}^n$ be the sequence of observations with
positive likelihood, i.e. $p(x_{z}^n)>0$. Then, for every $t$ such
that $z_2\leq z_1\leq 1\leq t\leq n$,
\begin{equation}\label{normihinnang}
\big\|\pi_t[x^n_{z_1}]-\pi_{t}[x_{z_2}^n]\big\|\leq
2\rho^{\kappa(x_1^t)},
\end{equation}
where $\rho\in(0,1)$ is as in (\ref{25}) and
$$j(t):=\lfloor {t-2\over r}\rfloor ,\quad\quad \kappa(x_1^t):=\sum_{u=0}^{j(t)-1}I_{{\cal
 X}_o^r}\big(x_{ur+2}^{(u+1)r+1}\big).$$
\end{lemma}
\begin{proof} Recall that for two transition matrices $A,B$,
$\delta(AB)\leq \delta(A)\delta(B)$, so
\begin{align*}
\delta\Big(\prod_{i=1}^{t-1}F_i\Big)=\delta\Big(\prod_{u=0}^{j-1}\Big(\prod_{i=ur+1}^{(u+1)r}F_i\Big)\prod_{i=jr+1}^{t-1}F_i\Big)\leq
\prod_{u=0}^{j-1}\delta\big(\prod_{i=ur+1}^{(u+1)r}F_i\big)=\prod_{u=0}^{j-1}\delta(A_u),\end{align*}
where $A_u:=\prod_{i=ur+1}^{ur+r}F_k[x_{ur+2}^n].$ From Proposition
\ref{clust-A}, with $k=ur+1$,
$$\delta(A_u)\leq \left\{
                  \begin{array}{ll}
                    \rho, & \hbox{if $x_{(ur+1)+1}^{(ur+1)+r}\in {\cal X}_o^r$;} \\
                    1, & \hbox{else.}
                  \end{array}
                \right.$$
From (\ref{hinnang2}), it holds
$$\big\|\pi_t[x^n_{z_1}]-\pi_{t}[x_{z_2}^n]\big\|\leq 2\delta\Big(\prod_{i=1}^{t-1}F_i[x_{i+1}^n]\Big) \leq
2\prod_{u=0}^{j-1}\delta(A_u)\leq 2\rho^{\kappa(x_1^t)}.$$
\end{proof}
\noindent
Let $s_1\in C$. By irreducibility and cluster assumption, there is a
path $s_1,\ldots, s_{r+1}$ such that $s_i\in C$ and
$P(Y_1=s_1,\ldots, Y_{r+1}=s_{r+1})>0$. By (i), for any
$s_2,\ldots,s_{r+1}\in C$, it holds $P(X_2^{r+1}\in
\X^{r}|Y_2=s_2,\ldots,Y_{r+1}=s_{r+1})>0$ implying that
$P(X_2^{r+1}\in \X^r)>0$. By stationarity of $X$, for every $k\geq
0$, it holds $P(X_1^{r}\in \X^r)=P(X_{k+1}^{k+r}\in \X^r)=:p_r>0.$
The process $\{X_n\}_{n\geq 1}$ is ergodic, so
\begin{equation}\label{12}
\lim_{t\to \infty}{\kappa(X_1^{t})\over t}=\lim_{t\to \infty}{1\over
r}{\kappa(X_1^{t})\over j(t)}={p_r\over r}>0,\quad {\rm
a.s.}.\end{equation}
\begin{corollary}\label{cor1} Assume {\bf A}. Then,
 there exists a  non-negative finite random variable $C_1$ as well as constant $\rho_1\in
(0,1)$ such that for every $z,t,n$ such that $z_2\leq z_1\leq 1\leq
t\leq n$,
\begin{equation}\label{random2}
\|P\big(Y_{t}\in \cdot|X_{z_1}^n\big)-P\big(Y_{t}\in \cdot
|X_{z_2}^n\big)\|\leq C_1{\rho_1}^{t-1},\quad \rm {a.s.}.
\end{equation}
\end{corollary}
\begin{proof}
The right hand side of (\ref{normihinnang}) does not depend on $n$
(as soon as it is bigger than $t$), hence from Lemma \ref{lemmake}
\begin{equation*}\label{random1a}
\sup_{n\geq t}\|P(Y_t\in \cdot |X_{z_1}^n)-P(Y_t\in \cdot
|X_{z_2}^n)\|\leq 2\rho^{\kappa(X_1^t)},\quad a.s..
\end{equation*}
Thus, if $t\to \infty$ then by (\ref{12}), it holds
\begin{equation}\label{limsup}
\limsup_{t}{1\over t}\log \Big(\sup_{n\geq t}\|P(Y_{t}\in \cdot
|X_{z_1}^n)-P(Y_{t}\in \cdot |X_{z_2}^n)\|\Big)\leq {p_r\over r}\log
\rho,\quad \rm{a.s.}.\end{equation}
Let $\varrho$  be such that  ${p_r\over r}\log \rho<\varrho <0.$ Let
\begin{equation}\label{T}
T(\omega):=\max\big\{ t\geq 1: {\log 2+\kappa(X_1^t)\log {\rho}\over t} > \varrho \big \}.
\end{equation}
From (\ref{limsup}), it follows that for almost every $\omega$,
$T(\omega)<\infty$ and for $t>T(\omega)$,
$$
\log \Big(\sup_{n\geq t}\|P\big(Y_{t}\in \cdot
|X_{z_1}^n\big)(\omega)-P\big(Y_{t}\in \cdot
|X_{z_2}^n\big)(\omega)\|\Big)\leq t \varrho $$ and, hence, for $t>T(\omega)$ and $n\geq t$,
$$\|P\big(Y_{t}\in \cdot
|X_{z_1}^n\big)-P\big(Y_{t}\in \cdot |X_{z_2}^n\big)\|\leq
e^{t\varrho}=({\rho}_1)^{t},$$ where
${\rho}_1:=e^{\varrho}$. The inequality (\ref{random2}) holds with
$C_1:=2\rho^{-T+1}$.\end{proof}
\noindent The inequality (\ref{limsup}) is similar to Theorem 2.2 in
\citep{LeGlandMevel}. The forgetting  equation in form
(\ref{random2}) is used in \citep{ungarlased}.
\begin{corollary}\label{cor2}
There exist a constant $\rho_1\in (0,1)$ and an ergodic process
$\{C_z\}_{-\infty}^{\infty}$ so that for any $z_2\leq z_1\leq z\leq
t \leq n$
\begin{equation}\label{random2z}
\|P\big(Y_{t}\in \cdot|X_{z_1}^n\big)-P\big(Y_{t}\in \cdot
|X_{z_2}^n\big)\|\leq C_z{\rho_1}^{t-z},\quad \rm {a.s.}.
\end{equation}
\end{corollary}
\begin{proof} The existence of $C_z$ follows exactly as in case $z=1$. The ergodicity of $\{C_z\}$ follows from the fact that the random
variables  $\{C_z\}$ are stationary coding of the ergodic process
$X$.\end{proof}
\begin{theorem}\label{main}
Assume {\bf A}. Then there exist a constant $\rho_o\in (0,1)$ and an
finite ergodic process $\{C'_z\}_{z=-\infty}^{\infty}$ so that for
every $z\leq 1\leq t\leq k\leq n$
\begin{equation}\label{thm}
 \|P(Y_t\in
\cdot|X_z^{n})-P(Y_t\in \cdot|X_{-\infty}^{\infty})\|\leq
C_1\rho_o^{t-1}+C'_k\rho_o^{k-t}\quad{\rm a.s.},\end{equation} where $C_1$ is  a finite random variable as in Corollary \ref{cor2}.
\end{theorem}
\begin{proof}
We reverse the time by defining $Y'_k=Y_{-k}$, $X'_k=X_{-k}.$ Thus,
$P(Y'_{-t}\in \cdot|{X'}_{-n}^{-z})=P(Y_{t}\in \cdot|{X}_{z}^{n}).$
It is easy to see that when HMM $(Y,X)$ satisfies {\bf A}, then so
does the reversed HMM $(Y',X')$. From Corollary \ref{cor2}, it
follows that there exists $\rho_2\in (0,1)$ and ergodic process
$\{C^{''}_{-z}\}$ so that for any $-n_2\leq -n_1 \leq -k\leq  -t
\leq -1 \leq -z$
$$\|P(Y_t\in \cdot|X_{z}^{n_2})-P(Y_{t}\in
\cdot|X_{z}^{n_1})\|=\|P(Y'_{-t}\in
\cdot|{X'}_{-n_2}^{-z})-P(Y'_{-t}\in \cdot|{X'}_{-n_1}^{-z})\|\leq
C^{''}_{-k}\rho_2^{-t-(-k)}=
C^{''}_{-k}\rho_2^{k-t}=C'_k\rho_2^{k-t},\quad {\rm a.s}$$ where
$C'_z:=C^{''}_{-z}$. The right side does not depend on $n_1$, $n_2$
and $z$. Hence, letting $z\to -\infty$ and using Levy martingale
convergence theorem, for every $1\leq t\leq k\leq n_1\leq n_2$
\begin{equation*}
\|P(Y_t\in \cdot|X_{-\infty}^{n_2})-P(Y_{t}\in
\cdot|X_{-\infty}^{n_1})\|\leq C'_k\rho_2^{k-t},\quad {\rm
a.s.}.\end{equation*} Letting now $n_1\to \infty$ and using Levy
martingale convergence theorem again, for every $1\leq t\leq k\leq
n$
\begin{equation}\label{eq:m}
\|P(Y_t\in \cdot|X_{-\infty}^{n})-P(Y_{t}\in
\cdot|X_{-\infty}^{\infty})\|\leq C'_k\rho_2^{k-t},\quad {\rm
a.s.}.\end{equation} Applying the same theorem to (\ref{random2}),
with $z_2\to -\infty$ and  $z=z_1$, we get that for every $z\leq
1\leq t \leq n$,
\begin{equation}\label{eq:z}
\|P(Y_t\in \cdot
|X_{-\infty}^n)-P(Y_t\in \cdot |X_{z}^n)\|\leq C_1\rho_1^{t-1}\quad
\rm {a.s.}.
\end{equation}
Hence, with, $\rho_o=\max\{\rho_1,\rho_2\}$, from the inequalities
(\ref{eq:m}) and (\ref{eq:z}), the inequality (\ref{thm}) follows.
 \end{proof}
\section{Convergence of risks}
\noindent Recall that $l: S\times S \to [0,\infty)$ is the pointwise
loss. Let, for any $1\leq t \leq n$, and $s\in S$,
$$R_t(s|x^n)=E[l(Y_t,s_t)|X^n=x^n]=\sum_{a\in S}
l(a,s)P(Y_t=a|X^n=x^n).$$ Thus, $R_t(s|x^n)$ is the conditional risk
of classifying $Y_t=s$ given the observations $x^n$. The risk of the
whole state sequence $s^n$ as defined in (\ref{risk}) with $L$ as in
(\ref{point-loss}) is easily seen to be
$$R(s^n|x^n)={1\over n}\sum_{i=1}^nR_t(s_t|x^n).$$
Let for every $t\in \mathbb{Z}$ and $s\in S$,
$$R_t(s|X_{-\infty}^{\infty}):=E[l(Y_t,s)|X_{-\infty}^{\infty}]=\sum_{a\in
S}l(a,s)P(Y_t=a|X_{-\infty}^{\infty}).$$ For $t\geq 1$, thus
\begin{equation}\label{riskivahe}
|R_t(s|X_{-\infty}^{\infty})-R_t(s|X_{1}^{n})|\leq l(s)\|P(Y_t\in
\cdot |X_{1}^n)-P(Y_t\in \cdot
|X_{-\infty}^{\infty})\|,\end{equation} where $l(s)=\max_al(a,s)$.
Finally, recall that $R(x^n):=\min_{s^n}R(s^n|x^n)$.
\begin{theorem}\label{riskthm}
Suppose  {\bf A} holds. Then there exists a constant $R$ such that
$R(X^n)\to R$ a.s. and in $L_1$.
\end{theorem}
\begin{proof}
The  process $X$ is ergodic, so for  a
constant $R$,
\begin{equation}\label{birkhoff}
{1\over n}\sum_{t=1}^n\min_sR_t(s|X_{-\infty}^{\infty})\to R,\quad
\text{ a.s. and in }L_1.\end{equation}
Let $M<\infty$ be such that $P(C'_n\leq M)=:q>0$. Let, for every
$n$, $k(n)=\max\{k\leq n: C'_k\leq M\}$. Since the process $C'$ is
ergodic, in the process $n\to\infty$, $k(n)\to \infty$, a.s. From
(\ref{riskivahe}), it follows, that with $A:=\max_{a,s} l(a,s)$,
$$|\min_sR_t(s|X_{-\infty}^{\infty})-\min_{s}R_t(s|X_{1}^{n})|\leq
A\|P(Y_t\in \cdot |X_{1}^n)-P(Y_t\in \cdot
|X_{-\infty}^{\infty})\|.$$ Hence
\begin{equation}\label{marta}
|R(X^n)-{1\over n}\sum_{t=1}^n \min_{s\in S}
R_t(s|X_{-\infty}^{\infty})|\leq  {A\over n}\sum_{t=1}^n\|P(Y_t\in
\cdot |X_{1}^n)-P(Y_t\in \cdot |X_{-\infty}^{\infty})\|\leq {A\over
n}\sum_{t=1}^{k(n)}\|P(Y_t\in \cdot |X_{1}^n)-P(Y_t\in \cdot
|X_{-\infty}^{\infty})\|+{A\over n}2(n-k(n)).
\end{equation}
 By inequality
(\ref{thm}), for every $1\leq t\leq k(n)$
$$
\sum_{t=1}^{k(n)}\|P(Y_t\in \cdot |X_{1}^n)-P(Y_t\in \cdot
|X_{-\infty}^{\infty})\|\leq
C_1\sum_{t=1}^{k(n)}\rho_o^{t-1}+M\sum_{t=1}^{k(n)}\rho_o^{k(n)-t}\leq
(C_1+M)\sum_{n=0}^{\infty}\rho_o^n<\infty,\quad \text{a.s.}.
$$
Let $\tau_1:=\min\{i\geq 0: C'_i\leq M\},\quad  \tau_j:=\min\{ i
>\tau_{j-1}: C'_i\leq M\}$.
Since $\{C'_k\}$ is ergodic, the random variables
$T_j=\tau_{j+1}-\tau_j$, $j=1,2,\ldots$ are identically distributed.
By Kac' return time theorem, $ET_j =q^{-1}$. Finally, denote
\begin{equation*}\label{j}
j(n)=\max\{j: \tau_j\leq n\}.
\end{equation*}
Thus $k(n)=\tau_{j(n)}$ and $n-k(n)\leq T_{j(n)}$. Since $T_j$ is
a.s. finite, clearly $j(n),k(n)\to \infty$ as $n$ grows. From the
finite expectation of $ET_j$, it follows that
$${T_j\over j}\to 0,\quad  {\rm a.s. },$$
implying that
\begin{equation}\label{un}
{n-k(n)\over n}\leq  {T_{j(n)}\over j(n)}\to 0,\quad \text{a.s.}
\end{equation}
Hence, the right hand side of (\ref{marta}) goes to 0, a.s. and from
(\ref{birkhoff}), it now follows that $R(X^n)\to R$ a.s. Risks are
nonnegative, so the convergence in $L_1$ follows from Sheffe's
lemma. \end{proof}
\noindent
Given $l$, the constant $R$ -- {\it asymptotic risk} -- depends  on
the model, only. It measures the average loss of classifying one
symbol using the optimal classifier.  For example, if $l$ is
symmetric, then the optimal classifier (in the sense of
misclassification error) makes in average about $Rn$ classification
errors. Clearly this is the lower bound: no other classifier does
better. The constant $R$ might be hard to determine theoretically,
but Theorem \ref{riskthm} guarantees that it can be approximated by
simulations.

\bibliographystyle{elsarticle-harv}
\bibliography{kirj}







\end{document}